\documentclass[11pt]{article}
\usepackage[preprint,nonatbib]{neurips_2026}

\usepackage{amsmath}
\usepackage{amssymb}
\usepackage{graphicx}
\usepackage{booktabs}
\usepackage{multirow}
\usepackage{hyperref}
\usepackage{float}
\title{
\textbf{Momentum-Guided Semantic Forecasting (MoFore) for Self-Supervised Video Representation Learning}
}

\author{
Qinwu Xu, PhD \thanks{This work was independently developed during Early Feb --Mid April 2026.} 
\\
\vspace{0.3cm} \\
\texttt{qinwu.xu2020@gmail.com}
}

\date{}

\begin{document}

\maketitle
\begin{abstract}

Self-supervised video representation learning has recently advanced through contrastive learning, masked reconstruction, and predictive representation learning. Reconstruction-based approaches such as MAE and VideoMAE learn representations by recovering masked visual content \cite{he2022mae,tong2022videomae}, while contrastive methods such as CLIP learn semantically meaningful embedding spaces through representation alignment \cite{radford2021clip}.

In this work, we introduce a Momentum-Guided Semantic Forecasting framework (MoFore) for self-supervised video representation learning. Instead of optimizing for pixel-level reconstruction or task-specific semantic alignment, the proposed method learns temporally predictive video representations by forecasting future latent embeddings from temporally distant context clips. To improve robustness across temporal scales, we further introduce randomized temporal-gap forecasting during training. The framework combines predictive latent forecasting with contrastive regularization to encourage temporal consistency while preventing representation collapse.

Experiments on the UCF101 dataset demonstrate that the proposed framework learns temporally consistent and semantically meaningful video representations without using action labels during training. Quantitative analysis shows strong temporal stability and emergent category-level structure in the learned embedding space, while qualitative retrieval experiments reveal motion-aware organization across related activities. Overall, the results suggest that long-range latent forecasting provides an effective and computationally efficient approach for self-supervised video representation learning without relying on reconstruction-based objectives.

\begin{figure}[!t]
    \centering
    \includegraphics[width=1\linewidth]{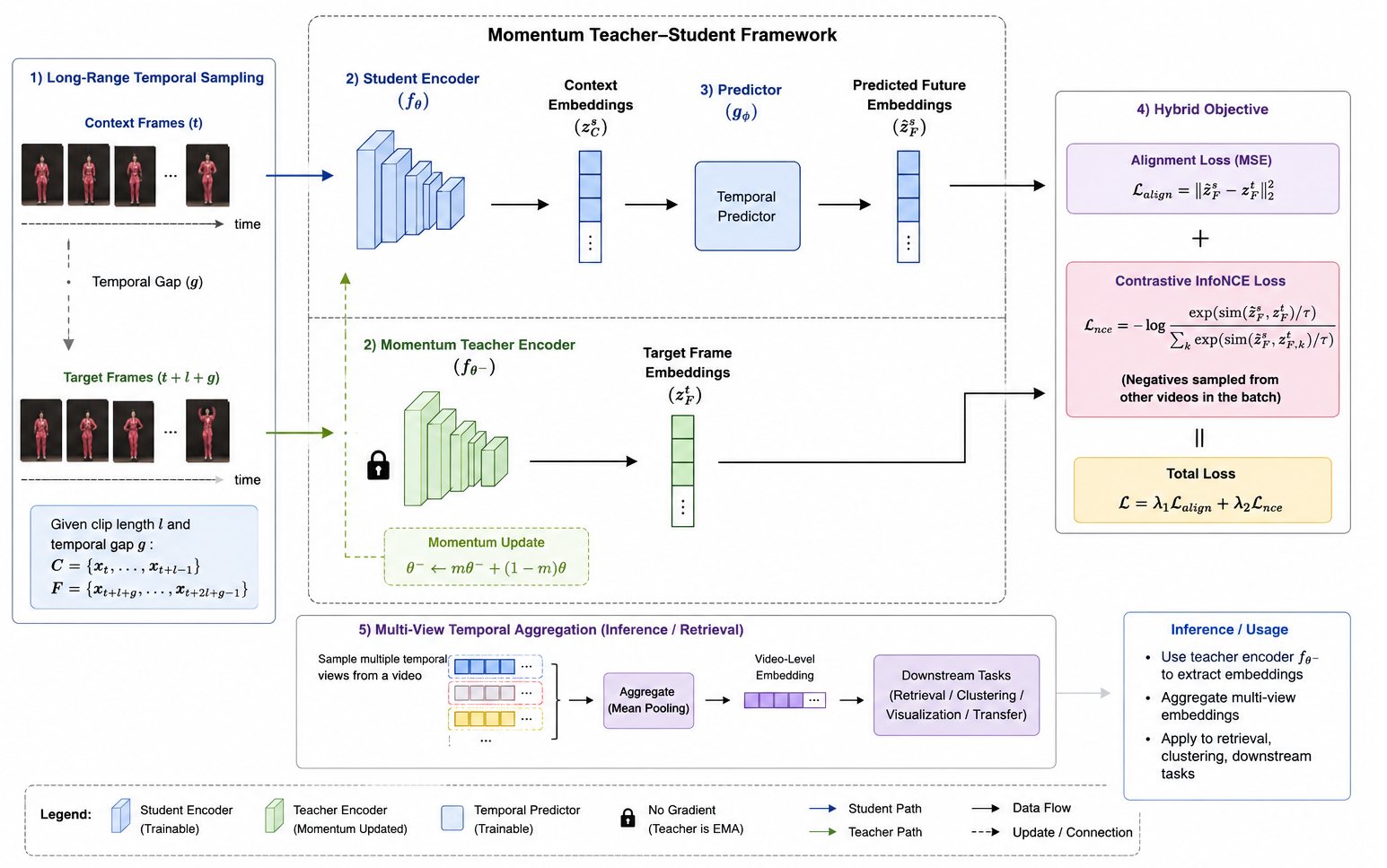}
    \caption{Architecture of the proposed momentum-guided long-range predictive learning framework.}
    \label{fig:architecture}
\end{figure}

\end{abstract}

\section{Introduction}

Learning semantic video representations without labels remains a fundamental challenge in computer vision. Unlike static images, videos contain rich temporal dynamics, motion patterns, and long-range dependencies that unfold over time. Effective video representations therefore require not only spatial understanding of individual frames but also the ability to model future semantic evolution across extended temporal horizons.

Recent advances in self-supervised learning have achieved remarkable success through contrastive learning, self-distillation, masked reconstruction, and latent predictive modeling \cite{chen2020simclr,he2020moco,grill2020byol,caron2021dino,he2022mae,tong2022videomae,lecun2022path}. Reconstruction-based approaches such as MAE \cite{he2022mae} and VideoMAE \cite{tong2022videomae} learn powerful visual representations by recovering masked image patches or video tokens, building on transformer-based backbones for vision and video \cite{dosovitskiy2021vit,bertasius2021space}. While highly effective, reconstruction objectives often require models to allocate substantial capacity toward recovering low-level appearance details that may not be directly relevant to high-level semantic understanding.

An alternative direction is predictive representation learning, which seeks to model the semantic structure directly in latent space rather than reconstructing pixels. Prior work including Contrastive Predictive Coding (CPC) \cite{oord2018cpc} and Joint Embedding Predictive Architectures (JEPA) \cite{lecun2022path} demonstrated that meaningful representations can emerge through predictive consistency between latent representations. CPC learns representations through autoregressive predictive learning. For example, given several visible image regions or patches, a context encoder summarizes the observed spatial context and predicts the latent representation of neighboring unseen patches. The model is trained contrastingly to distinguish the true latent representation from negative samples drawn from other image regions or images, encouraging discriminative predictive representation learning. In contrast, JEPA focuses on predicting masked spatial representations within a joint embedding space, typically emphasizing latent prediction of larger masked image regions without explicit contrastive discrimination. 

Our work instead focuses on long-range semantic forecasting for video representation learning using momentum-guided latent prediction. Rather than autoregressively predicting nearby latent states with contrastive discrimination, the proposed framework learns to forecast temporally distant video-level latent representations through direct latent-space alignment between a student network and a momentum-updated teacher network. Entire future temporal segments remain unobserved during prediction, encouraging the model to capture temporally persistent semantic structure and motion dynamics across substantial temporal separation. To improve robustness across different temporal scales, we further introduce randomized temporal-gap sampling during training, exposing the model to diverse forecasting horizons.

 The proposed framework combines semantic forecasting objectives with contrastive regularization to simultaneously preserve temporal consistency and maintain embedding diversity. Through long-range predictive learning, semantic structure emerges naturally from forecasting future video dynamics rather than recovering visual appearance.

Our experimental results on UCF101 demonstrate that the learned representations exhibit meaningful semantic clustering, strong temporal consistency, and motion-aware retrieval behavior despite the absence of action labels during training. These findings suggest that long-range semantic forecasting provides an effective and computationally efficient paradigm for self-supervised video representation learning.

The main contributions of this work are summarized as follows:

\begin{enumerate}
\item We introduce a semantic forecasting framework for self-supervised video representation learning that learns temporally predictive latent representations without relying on pixel reconstruction or task-specific supervised objectives.

\item We propose randomized long-range temporal-gap forecasting, enabling representation learning across diverse temporal horizons and improving robustness to varying temporal scales.

\item We develop a momentum-guided latent forecasting objective in which a momentum-updated teacher network provides stable future representation targets for long-range temporal prediction.

\item Experiments on UCF101 demonstrate that the proposed framework learns temporally consistent and semantically organized video representations without using action labels during training.
\end{enumerate}

\section{Related Work}

\subsection{Contrastive and Self-Distillation Representation Learning}

Self-supervised representation learning has achieved substantial progress through contrastive and self-distillation objectives. Contrastive learning methods such as SimCLR \cite{chen2020simclr} and MoCo \cite{he2020moco} learn visual representations by maximizing agreement between positive pairs while separating negative samples. These approaches demonstrated that strong semantic representations can emerge from large-scale self-supervised objectives without requiring manual annotations.

Subsequent methods such as BYOL \cite{grill2020byol} and DINO \cite{caron2021dino} further showed that meaningful representations can be learned without explicit negative pairs through momentum teacher architectures and self-distillation objectives. These methods highlighted the importance of predictive consistency, representation stability, and momentum-based target generation for effective self-supervised learning.

\subsection{Reconstruction-Based Self-Supervised Learning}

Reconstruction-based methods learn visual representations by recovering masked observations. MAE \cite{he2022mae} reconstructs masked image patches using Vision Transformers \cite{dosovitskiy2021vit}, while VideoMAE \cite{tong2022videomae} extends this paradigm to video representation learning through masked spatiotemporal reconstruction, building on transformer-based video modeling \cite{bertasius2021space}.

These approaches have demonstrated strong downstream performance across a variety of visual tasks. However, reconstruction objectives may require substantial modeling capacity to recover low-level appearance details, textures, and local visual statistics that are not always essential for high-level semantic understanding. This observation has motivated growing interest in learning semantic representations directly in latent space without explicit pixel reconstruction.

\subsection{Predictive Representation Learning and Semantic Alignment}

An alternative paradigm in self-supervised learning is predictive representation learning, which seeks to model future semantic structure directly in a latent space rather than reconstructing high-dimensional pixel data. Early work such as Contrastive Predictive Coding (CPC) \cite{oord2018cpc} demonstrated that meaningful abstractions can emerge through future latent prediction objectives. More recently, the Joint Embedding Predictive Architecture (JEPA) \cite{lecun2022path} proposed predictive world modeling via latent-space target prediction[cite: 1], while VL-JEPA \cite{chen2025vljepa} extended this paradigm to multimodal vision-language settings. 

Concurrently, advances in large-scale representation alignment—including dual-encoder frameworks like CLIP \cite{radford2021clip}, MetaCLIP \cite{xu2023metaclip}, and SigLIP \cite{zhai2023sigmoid}, alongside multimodal architectures such as Flamingo \cite{alayrac2022flamingo}, BLIP-2 \cite{li2023blip2}, and LLaVA \cite{liu2023llava}—have demonstrated that robust visual embedding spaces naturally emerge through cross-modal semantic alignment. Recent literature has further highlighted the critical role of semantic consistency, latent-space alignment, and representation stability in enhancing the robustness of complex multimodal learning systems (\cite{xu2026}, \cite{xu2026ocr}, and \cite{xu2026checkpoint}). 

Crucially, while some recent predictive frameworks focus heavily on dense, pixel-level future semantic forecasting—such as FUTURIST \cite{karypidis2025advancing}, which utilizes masked visual transformers to generate explicit, high-resolution future semantic segmentation masks and depth maps \cite{karypidis2025advancing}—our proposed method differs fundamentally (see Table~\ref{tab:method_comparison}). Instead of generating explicit, high-dimensional downstream maps or relying on localized visual tokens, our Momentum-Guided Semantic Forecasting (MoFore) framework optimizes for global, abstract video-level representation learning directly within the latent space. This structural distinction allows our model to capture long-range action dynamics without incurring the heavy computational and parametric overhead typically associated with dense autoregressive token generation.

\subsection{Self-Supervised Video Representation Learning}

Self-supervised video learning methods seek to model temporal structure and motion dynamics without human labels by leveraging temporal ordering, contrastive learning, masked reconstruction, or predictive objectives \cite{he2022mae,tong2022videomae,lecun2022path,bertasius2021space,oord2018cpc,chen2025vljepa,wang2021temporal}. Early approaches utilized autoregressive future latent prediction \cite{oord2018cpc} or exploited the local temporal coherence between neighboring video clips through temporal contrastive learning \cite{wang2021temporal}. Conversely, recent reconstruction-based paradigms, such as VideoMAE \cite{tong2022videomae}, have achieved strong performance by training Vision Transformers to recover highly masked spatiotemporal visual tokens \cite{feichtenhofer2022masked}. While effective, such reconstruction objectives often force a backbone to dedicate substantial modeling capacity to low-level appearance details and local frame continuity rather than high-level semantic progression.

To address this limitation, the proposed framework explicitly formulates video representation learning as a long-range semantic forecasting problem. Rather than focusing on pixel reconstruction or short-range visual continuity, we forecast future semantic states across randomized long-range temporal horizons using stable, momentum-guided targets. This strategy forces the learned embedding space to remain sensitive to temporally persistent semantic structures and broad motion dynamics. This design aligns with recent insights showing that preserving semantic consistency within latent representations significantly improves structural robustness under distribution shifts \cite{xu2026miller}. 

To clearly contextualize our positioning within the literature, we provide a structural and conceptual comparison against foundational paradigms in Table~\ref{tab:method_comparison}. Ultimately, while our framework shares the stable momentum-updating paradigm of MoCo \cite{he2020moco} and aligns with the core philosophy of FUTURIST \cite{karypidis2025advancing} by predicting high-level semantics over raw visual pixels, it uniquely bridges the gap between these paradigms by optimizing specifically for abstract, long-range latent temporal representation learning.

\begin{table*}[t]
\centering
\caption{Structural and conceptual comparison between MoCo, FUTURIST, and our proposed framework (MoFore).}
\label{tab:method_comparison}
\resizebox{\textwidth}{!}{%
\begin{tabular}{l|lll}
\hline
\textbf{Dimension} & \textbf{MoCo} (He et al., 2020) & \textbf{FUTURIST} (Karypidis et al., 2025) & \textbf{Our Method (MoFore)} \\ \hline
\textbf{Core Task} & Self-supervised frame instance discrimination & Dense, pixel-level future scene forecasting & Abstract, long-range latent temporal forecasting \\
\textbf{Output Space} & Low-dimensional global instance embeddings & High-resolution dense maps (Segmentation/Depth) & Aggregated, motion-aware video-level latent vectors \\
\textbf{Target Split} & Spatial augmentations of the \textit{same} static frame & Sequentially continuous future frames & Temporally distant context vs. target clip pairs \\
\textbf{Architecture} & Siamese dual-encoder with a dynamic queue & Spatio-temporal masked visual transformer & Student-teacher frame aggregator with a predictor \\
\textbf{Loss Function} & InfoNCE contrastive loss & Per-modality masked cross-entropy loss & Hybrid Alignment (MSE) + Contrastive Regularization \\ \hline
\end{tabular}%
}
\end{table*}

\section{Method}

\subsection{Overview}
Figure~\ref{fig:architecture} illustrates the overall architecture of the proposed Momentum-Guided Semantic Forecasting (MoFore) framework. Given an unlabeled video sequence

\begin{equation}
V = \{x_1, x_2, \ldots, x_T\}
\end{equation}

the objective is to learn semantic video representations through long-range temporal forecasting. Rather than reconstructing future frames or recovering masked visual tokens, the proposed framework learns to forecast future semantic states directly in latent representation space.

The framework consists of three components: (1) a student encoder that extracts context representations, (2) a momentum-updated teacher encoder that generates future semantic targets, and (3) a lightweight forecasting network that predicts future latent states from temporally separated context clips.

Given a context clip and a future target clip sampled from the same video sequence, the student network learns to forecast the semantic representation of the future clip, while the teacher network provides stable forecasting targets. Through repeated long-range semantic forecasting across diverse temporal horizons, the model learns representations that capture both appearance and temporal evolution.

Unlike reconstruction-based approaches that focus on recovering visual details, the proposed framework directly models future semantic consistency in latent space. This formulation encourages representations that remain predictive under substantial temporal separation and emphasizes temporally persistent semantic structure rather than short-range visual continuity.

\subsection{Long-Range Temporal Semantic Forecasting}

A central design principle of the proposed framework is that meaningful video semantics often emerge over extended temporal intervals. Predicting adjacent frames may be solved using local appearance cues, whereas forecasting temporally distant semantic states requires understanding longer-range motion dynamics and action evolution.
\begin{figure}
    \centering
    \includegraphics[width=1\linewidth]{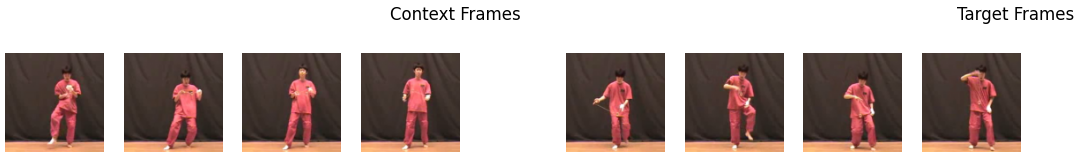}
    \caption{Temporal forecasting illustration}
    \label{fig:context_target}
\end{figure}
Figure~\ref{fig:context_target} illustrates the temporal forecasting setup used during training. Given clip length $l$ and temporal gap $g$, a context clip and a future target clip are sampled as

\begin{equation}
C={x_t,\ldots,x_{t+l}},
\end{equation}

\begin{equation}
F={x_{t+l+g},\ldots,x_{t+2l+g}}.
\end{equation}

Unlike fixed-horizon prediction methods, the temporal gap $g$ is randomly sampled during training. This randomized temporal-gap forecasting strategy exposes the model to diverse forecasting horizons and prevents over-specialization to a particular temporal scale.

By requiring accurate forecasting across both short and long temporal separations, the learned representation becomes increasingly sensitive to temporally persistent semantic structure while remaining robust to local appearance variations.

\subsection{Momentum-Guided Semantic Forecasting Architecture}

Each video frame is first encoded using a backbone encoder

\begin{align}
z_i = E_\theta(x_i),
\end{align}

where

\[
z_i \in \mathbb{R}^d
\]

denotes the latent frame representation.

Frame-level features are temporally aggregated to obtain clip-level semantic representations. The student encoder processes the context clip and produces a context representation $z_c$.

To generate stable future semantic targets, a momentum teacher encoder processes the future clip:

\begin{align}
z_t = E_\xi(F).
\end{align}

The teacher parameters are updated using exponential moving average

\begin{align}
\xi \leftarrow m\xi + (1-m)\theta,
\end{align}

where $m$ denotes the momentum coefficient.

A lightweight forecasting network maps the context representation into the future semantic space:

\begin{align}
\hat{z}t=P\phi(z_c),
\end{align}

where

\begin{align}
P_\phi(z)=W_2\sigma(W_1z).
\end{align}

The forecasting objective encourages the predicted future semantic state $\hat{z}_t$ to align with the teacher-generated future target $z_t$. Through this process, the model learns latent representations that remain predictive across substantial temporal gaps.

Importantly, the teacher network is not used to reconstruct future observations. Instead, it provides semantic forecasting targets that guide representation learning toward temporally stable abstractions.

\subsection{Semantic Forecasting Objective}

A forecasting objective alone may lead to representational collapse. To encourage both predictive consistency and representation diversity, we combine semantic forecasting loss with contrastive regularization.

The semantic forecasting loss minimizes the discrepancy between the predicted future semantic state and the teacher-generated target:

\begin{equation}
\| \hat{z}_t - z_t \|^2
\end{equation}

Given a batch of positive pairs

\begin{align}
{(z_i,z_i^+)}_{i=1}^{B},
\end{align}

the contrastive regularization term is defined as
\begin{equation}
-\log
\frac{
\exp(\operatorname{sim}(z_i,z_i^+)/\tau)
}{
\sum_j
\exp(\operatorname{sim}(z_i,z_j)/\tau)
}.
\end{equation}

The overall training objective becomes
\begin{equation}
L_{\text{forecast}}
+
\lambda L_{\text{con}},
\end{equation}

where $\lambda$ controls the strength of contrastive regularization.

The forecasting term encourages long-range semantic consistency across time, while the contrastive component promotes discriminative and information-rich representations. Together, these objectives enable semantic video representations to emerge from forecasting future latent states rather than reconstructing visual observations.

\section{Experiments}

We evaluate whether long-range semantic forecasting can learn meaningful video representations without labels. Experiments were conducted on the UCF101 action recognition dataset \cite{soomro2012ucf101}, which contains 13,320 realistic YouTube video clips spanning 101 human action categories and approximately 27 hours of video data. The dataset covers diverse activity types including sports, body-motion activities, human-object interactions, and musical performance actions such as Basketball Dunk, YoYo, Playing Guitar, and Apply Eye Makeup. Videos are organized into 25 groups with shared visual characteristics such as similar backgrounds or viewpoints, enabling standardized train/test evaluation splits. The clips are short trimmed videos with an average duration of approximately 7 seconds, recorded at 25 FPS with substantial variation in camera motion, pose, viewpoint, illumination, and background clutter, making UCF101 a widely used benchmark for self-supervised video representation learning and action recognition research. Although category labels are available for evaluation, they are never used during training.

Videos were randomly divided into training and validation subsets using an 80/20 split. During training, context and target clips were sampled using randomized temporal windows with varying temporal separations. This setup encourages the model to learn representations that remain predictive across diverse forecasting horizons rather than relying on short-range visual continuity.

The proposed framework was trained using the Adam optimizer with learning rate $1\times10^{-4}$, batch size 8, and teacher momentum coefficient 0.7. Student encoder and forecasting network parameters were optimized through backpropagation, while teacher parameters were updated using exponential moving average.

To evaluate representation quality, we measure several embedding-level diagnostics on held-out validation videos. Specifically, we analyze (1) future semantic forecasting accuracy, (2) temporal consistency across video segments, (3) representation stability under multiple temporal views of the same video, and (4) semantic similarity relationships between videos from the same and different action categories. In addition, nearest-neighbor retrieval experiments are performed to qualitatively examine the semantic structure emerging in the learned representation space.

\section{Results}

\subsection{Quantitative Representation Analysis}

Table~\ref{tab:quantative} summarize representative quantitative results on the held-out validation split. The learned representations also exhibit strong temporal stability. Different temporal views sampled from the same video achieve near-perfect similarity (e.g., 0.9995 for different temporal clips from YoYo\_g19\_c02), while neighboring temporal representations within the same sequence maintain high temporal consistency (0.9223). These results indicate that the learned representation remains highly stable under temporal resampling and local motion variation, which is desirable for semantic forecasting because the underlying action semantics remain largely unchanged across different observations of the same video.

Beyond temporal stability, the learned embedding space exhibits emergent category-level semantic organization. Videos belonging to the same action category achieve substantially higher similarity (e.g., 0.7901 between YoYo\_g19\_c02 and YoYo\_g20\_c01) than videos from unrelated categories (e.g., 0.5521 between YoYo\_g19\_c02 and SkyDiving\_g06\_c03). This gap suggests that long-range semantic forecasting encourages semantically structured representations despite the complete absence of action labels during training.

The observed different-class similarity remains moderately high for this example. This behavior is expected because the videos still share coarse visual and kinematic characteristics, including human-centered motion, outdoor scenes, similar camera viewpoints, and large-scale body dynamics. Unlike supervised classification objectives that explicitly maximize inter-class separation, the proposed self-supervised forecasting objective primarily encourages predictive temporal consistency and semantic structure in latent space.

Overall, the quantitative results suggest that long-range semantic forecasting successfully produces temporally stable, semantically organized, and non-collapsed video representations. The low forecasting error, strong temporal consistency, and emergent category-level structure collectively support the effectiveness of semantic forecasting as a self-supervised video representation learning objective.

\begin{table}[H]
\centering
\caption{
Quantitative evaluation of the proposed Momentum-Guided Semantic Forecasting framework of UCF101 example (Class Yoyo - g19 \& g20 versus Class Skydiving). Metrics are computed on the held-out validation split using cosine similarity between normalized video representations.
}
\label{tab:quantative}
\begin{tabular}{p{4cm}p{8.2cm}c}
\toprule
Metric & Evaluation Protocol & Value \\
\midrule

Future Semantic Forecasting Error &
Mean squared error between predicted future semantic representations and momentum-teacher target representations sampled from temporally separated clips.
& 0.0039 \\

Same-Video Similarity &
Cosine similarity between representations extracted from different temporal views of the same video.
& 0.9995 \\

Same-Class Similarity &
Cosine similarity between representations extracted from different videos belonging to the same UCF101 action category.
& 0.7901 \\

Different-Class Similarity &
Cosine similarity between representations extracted from videos belonging to unrelated action categories.
& 0.5521 \\

Temporal Consistency &
Cosine similarity between neighboring temporal representations within the same video sequence.
& 0.9223 \\
Video example &\multicolumn{2}{c}{
\includegraphics[width=0.6\linewidth]{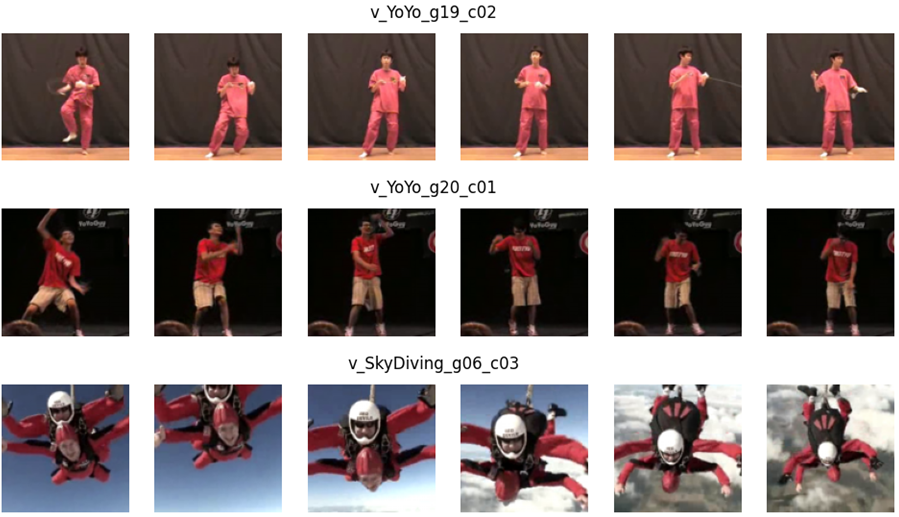}} \\
\bottomrule
\end{tabular}
\end{table}

\subsection{Qualitative Retrieval Analysis}

While quantitative metrics measure forecasting accuracy and representation consistency, retrieval experiments provide a more direct view of the semantic structure learned through long-range semantic forecasting. For each query video, nearest neighbors are retrieved using cosine similarity in the learned representation space.

\begin{figure}[t]
\centering
\includegraphics[width=1\linewidth]{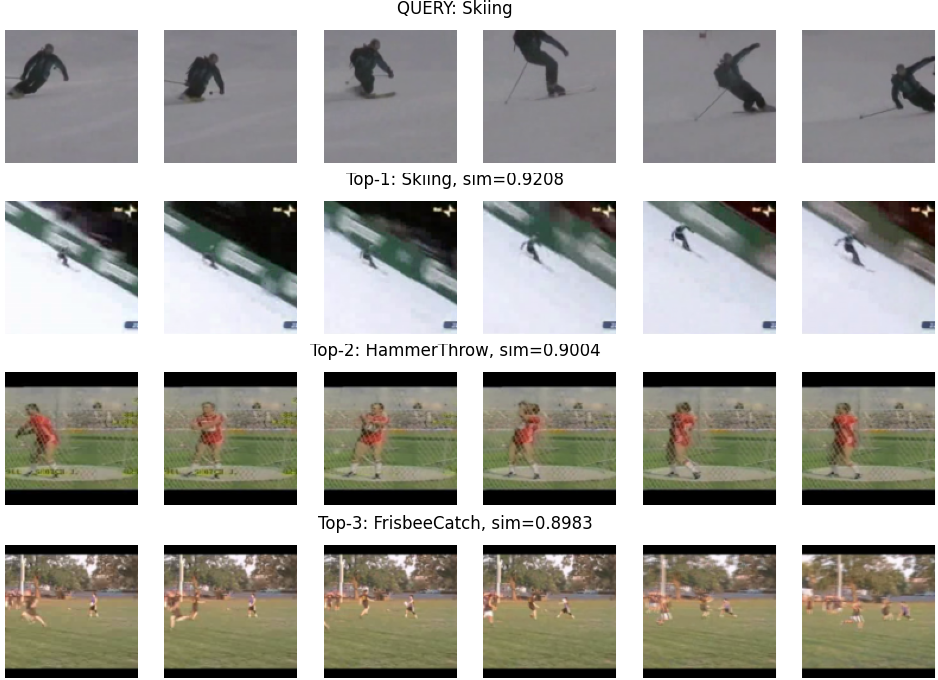}
\caption{Nearest-neighbor retrieval results for a Skiing query video. Retrieved videos exhibit similar motion dynamics, body posture, and athletic movement patterns, suggesting that long-range semantic forecasting captures high-level kinematic structure.}
\label{fig:Skiing}
\end{figure}

Figure~\ref{fig:Skiing} presents one of the strongest retrieval examples. Given a query video from the \textit{Skiing} category, the highest-ranked retrieval is another skiing sequence with cosine similarity 0.9208. The retrieved video exhibits highly similar body posture, directional motion, and environmental context, suggesting that the learned representation successfully captures characteristic skiing dynamics.

Interestingly, additional retrieved examples such as \textit{HammerThrow} and \textit{FrisbeeCatch} do not belong to the same semantic category but nevertheless share large-scale body rotations, coordinated limb movement, and athletic motion patterns. This behavior indicates that the representation space is organized not only by appearance but also by higher-level kinematic structure. Such retrieval patterns are consistent with the semantic forecasting objective, which encourages representations to capture motion dynamics that remain predictive across extended temporal horizons.

\begin{figure}[t]
\centering
\includegraphics[width=1\linewidth]{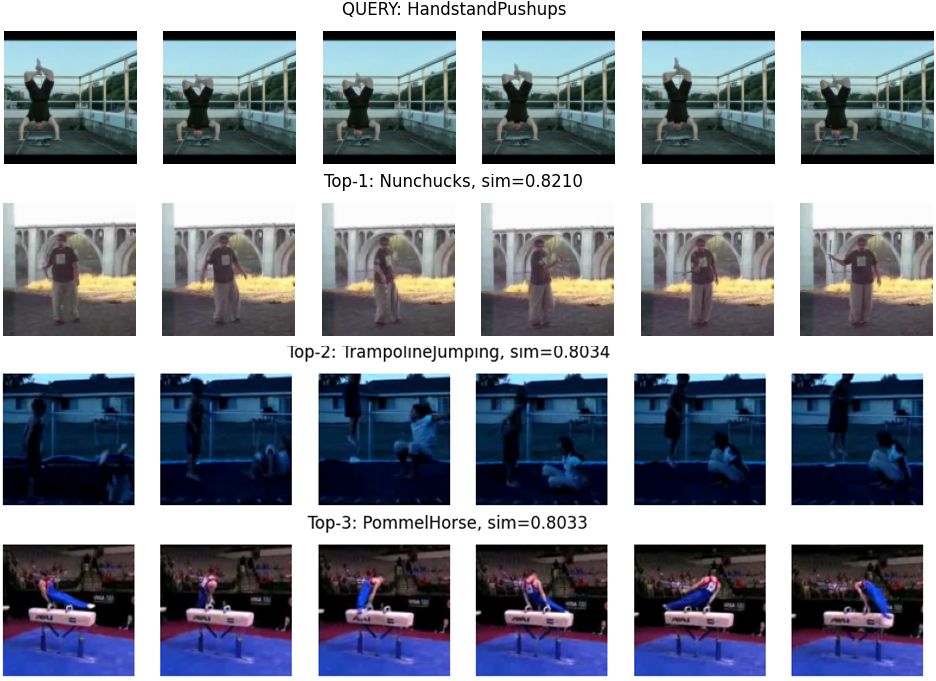}
\caption{Nearest-neighbor retrieval results for a HandstandPushups query video. Retrieved examples share similar body configurations and dynamic movement patterns despite belonging to different action categories.}
\label{fig:HandstandPushups}
\end{figure}

A similar phenomenon is observed in Figure~\ref{fig:HandstandPushups} \textit{HandstandPushups} category, where the retrieved results include \textit{Nunchucks}, \textit{TrampolineJumping}, and \textit{PommelHorse}. Although these activities belong to different semantic categories, they share distinctive body configurations involving inverted poses, strong vertical motion, and dynamic full-body movement.

The retrieval behavior suggests that the learned representation is sensitive to pose-level and motion-level structure without requiring explicit supervision. Rather than relying solely on scene appearance, the model appears to organize videos according to latent semantic factors that remain predictive of future motion evolution.

\begin{figure}[t]
\centering
\includegraphics[width=1\linewidth]{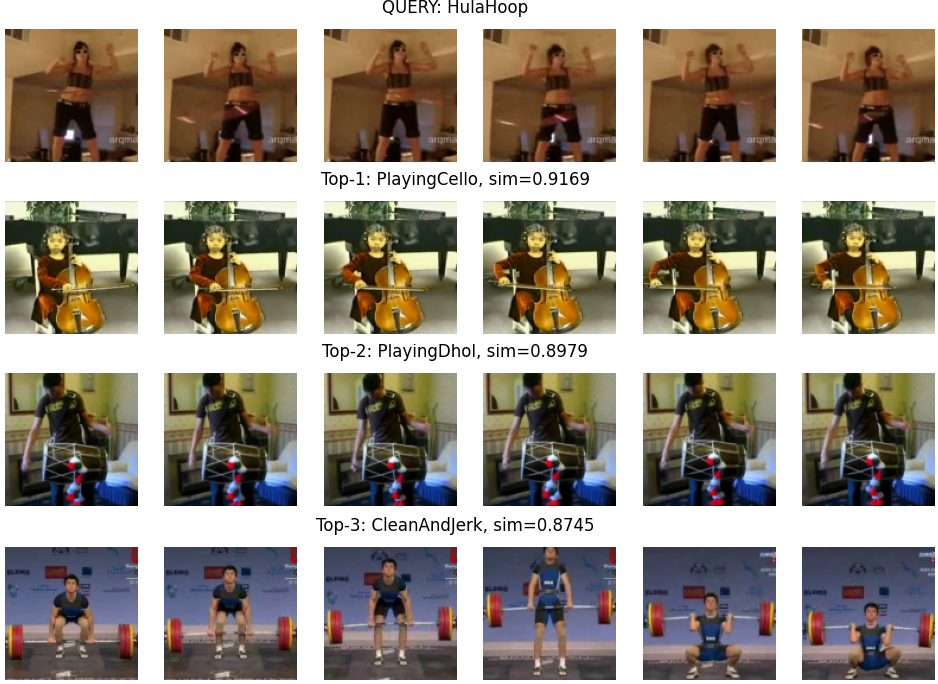}
\caption{Nearest-neighbor retrieval results for a Hula Hoop query video. Failure cases reveal that the current representation occasionally groups actions with similar pose geometry and repetitive motion patterns despite different semantic meanings.}
\label{fig:HulaHoop}
\end{figure}

The retrieval results also reveal some limitations of the current framework. Figure~\ref{fig:HulaHoop} shows that a query video from the \textit{Hula Hoop} category retrieves activities of \textit{Playing Cello} and \textit{Playing Dhol}. While semantically distinct, these activities share upright human posture, centered body positioning, and repetitive arm movement. The retrieval list also includes \textit{Clean and Jerk}, which involves substantially different action semantics but contains visually salient full-body motion patterns.

These examples suggest that the current representation remains strongly influenced by coarse pose geometry and overall motion structure. As a result, actions that differ semantically but exhibit similar visual configurations may still occupy nearby regions in the latent space.

These observations are consistent with the current architectural design. The present implementation primarily aggregates frame-level representations through temporal averaging and does not explicitly model fine-grained temporal ordering, motion trajectories, or long-range action progression. Consequently, the framework learns robust representations of pose, scene context, and coarse motion dynamics while still struggling with action categories that require detailed temporal reasoning.

Nevertheless, the retrieval results provide strong qualitative evidence that meaningful motion-aware latent structure emerges from long-range semantic forecasting. The learned representation space captures nontrivial relationships among actions and organizes videos according to shared kinematic patterns, pose evolution, and coarse motion trajectories despite the absence of action labels during training. Retrieved neighbors frequently exhibit similar body configurations and temporal dynamics even across different semantic categories, suggesting that long-range latent forecasting encourages representations that capture higher-level action dynamics beyond isolated appearance cues alone.

 \section{Conclusion}

We introduced a Momentum-Guided Semantic Forecasting framework (MoFore) for self-supervised video representation learning. Instead of reconstructing pixels or masked video content, the proposed approach learns video representations by forecasting future semantic states across temporally separated video segments in latent space.

The framework combines randomized long-range temporal forecasting, momentum-guided teacher supervision, and contrastive regularization within a lightweight student--teacher architecture. By learning to predict future semantic representations rather than future visual observations, the model encourages the emergence of temporally stable and semantically meaningful video representations.

Experimental results on UCF101 demonstrate that long-range semantic forecasting produces representations with strong temporal consistency, meaningful category-level organization, and motion-aware retrieval behavior despite the absence of action labels during training. Qualitative retrieval analysis further suggests that the learned representation space captures higher-level motion dynamics and semantic structure beyond simple appearance similarity.

Overall, the results support the central hypothesis of this work: forecasting future semantic states provides an effective alternative to reconstruction-based objectives for self-supervised video representation learning. The proposed framework offers a simple and computationally efficient approach for learning temporally predictive video representations and highlights semantic forecasting as a promising direction for future self-supervised video understanding research.

\subsection{Future Work}

Several directions may further extend the proposed semantic forecasting framework.

First, the current implementation relies on frame-level representations with a relatively simple temporal aggregation for latency consideration. Incorporating Transformer-based temporal encoders and forecasting modules may improve the modeling of long-range action dynamics and temporal dependencies.

Second, future work may investigate hierarchical semantic forecasting across multiple temporal horizons, enabling representations to capture both short-term motion patterns and long-term action evolution within a unified framework.

Third, larger-scale video pretraining datasets such as Kinetics, Something-Something, or Ego4D may provide richer temporal diversity and enable a more comprehensive evaluation of semantic forecasting at scale.

Finally, the semantic forecasting paradigm naturally extends to multimodal learning settings. Future research may explore forecasting future semantic states jointly across video, audio, language, and embodied interaction streams, potentially bridging self-supervised video representation learning and predictive world modeling.

\end{document}